# AUTONOMOUS FARM VEHICLES PROTOTYPE OF POWER REAPER

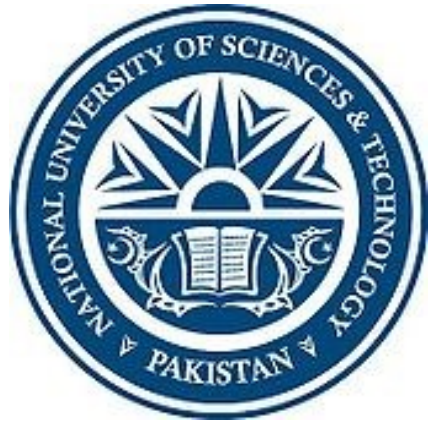

School Of Electrical Engineering and Computer Science

ADVISOR

Dr. Amir Ali Khan

COADVISOR

Dr. Muhammad Murtaza Khan

GROUP MEMBERS

| | |
|---|---|
| Abdul Qadeer Khan | 2009-NUST-BE-EL-02 |
| Ayyaz Akhtar | 2009-NUST-BE-EL-18 |
| Muhammad Zubair Ahmad | 2009-NUST-BE-EL-79 |

NUST School of Electrical Engineering and Computer Science

NUST H-12, Islamabad - 44000, Pakistan. Ph. No: 90852400

# Approval Page

This project report titled **"Autonomous Farm Vehicles- Prototype of a Power Reaper"** is submitted to fulfill the partial requirement of the degree of Bachelor of Electrical Engineering by **Abdul Qadeer Khan, Ayyaz Akhtar and Muhammad Zubair Ahmad** and is approved by the committee on: _____________________

X________________________________  
Dr. Amir Ali Khan  
Project Advisor

X________________________________  
Dr. Muhammad Murtaza Khan  
Projet Co- Advisor

# Dedication

To Allah who granted us the ability for this task.

To the Prophets who guided the humanity.

To the saints who spread message of love.

To our Country which has given us an identity.

To our Teachers who have taught us what we know.

To Family for their love and patience.

To Friends for their support.

To the Farmers who feed the nation with the sweat of their brow.

To those who dare to innovate

And

I (Muhammad Zubair) thank my group mates for dedicating this work to my uncle Riffat Ali Asghar who died in a Tractor accident, may his soul rest in peace

# Acknowledgements

We are sincerely thankful to our advisor, Dr. Amir Ali Khan, and our co-advisor, Dr. Muhammad Murtaza Khan, for their help and guidance during this project. We would like to give special thanks to Dr. John Billingsely (University of new Southern Queensland) for his guidance in the initial stages of the project. Beside them we would like to thank Dr. Sami-ur-Rehman and Col. Naveed from SMME for their help in mechanical design. Our parents and friends deserve special mention for their support.

# Table of Contents

# List of Tables

# List of Figures

*Chapter 1*

# INTRODUCTION

## 1.1 Problem Overview

Pakistan is primarily an agricultural country. This sector contributes 21% to GDP and employs 45% of the labor force (refer to Table 1: Industry wise employed labor). 60% of the rural population depends on it for their livelihood [1]. This sector is facing a lot of problems which are hindering its progress.

Major problems are:

- Non-availability of labor due to increased rural-urban migration due to increased industrialization. Industries provide better working conditions and higher wages than the agricultural sector and the industrial employment is generally not seasonal. The effect of the "industrial-pull" vary with the distance from the urban center. Migration of labor varies from 6-14% for landless agricultural labor and 7-33% for tenants [2].
- Extreme weather condition make it difficult for the labor to work efficiently and safely in the fields. Our major crop cover lies in Punjab and Sindh which are one of the hottest regions of the country with temperatures crossing 40 degrees in shade(refer to Figure 1: Pakistan Temperature map). Efficiency is the key in agricultural processes as rain can cause damage to both ripe and harvested crop. Sometimes harvesting is a race against time and weather.
- Excessive UV radiation (refer to Figure 2: UV index Pakistan showing the ammount of incident solar ultra voilet radiation) can cause skin cancer, cataracts and other related problems.
- Pakistan's average yield is 27.4 mans/acre. As the support price of wheat is Rs.1050/man so a farmer's revenue is Rs.28, 875/acre. They have to pay for the fertilizers, the ploughing, seeding, harvesting and threshing. They also have to save for the next crop which leaves them in deep financial troubles.

| Table-12.11: Employment Shares by Industry (%) | | | | | | | | | |
|---|---|---|---|---|---|---|---|---|---|
| Major Industry Divisions | 2008-09 | | | 2009-10 | | | 2010-11 | | |
| | Total | Male | Female | Total | Male | Female | Total | Male | Female |
| **Total** | **100** | **100** | **100** | **100** | **100** | **100** | **100** | **100** | **100** |
| Agriculture/ forestry/ hunting & fishing | 45.1 | 37.3 | 74.0 | 45.0 | 36.6 | 74.9 | 45.1 | 36.2 | 75.4 |
| Manufacturing | 13.0 | 13.3 | 11.9 | 13.2 | 13.9 | 11.0 | 13.7 | 14.5 | 10.9 |
| Construction | 6.6 | 8.3 | 0.4 | 6.7 | 8.5 | 0.3 | 7.0 | 8.9 | 0.2 |
| Wholesale & retail trade | 16.5 | 20.5 | 1.6 | 16.3 | 20.2 | 2.1 | 16.2 | 20.4 | 1.6 |
| Transport/ storage & communication | 5.2 | 6.6 | 0.2 | 5.2 | 6.6 | 0.3 | 5.1 | 6.6 | 0.1 |
| Community/social & personal service | 11.2 | 11.1 | 11.6 | 11.2 | 11.2 | 11.2 | 10.8 | 10.8 | 11.5 |
| *Others | 2.4 | 2.9 | 0.3 | 2.4 | 3.0 | 0.2 | 2.1 | 2.6 | 0.3 |

*Table 1: Industry wise employed labor showing that agriculture sector employees majority of the labor in Pakistan [1] [4]*

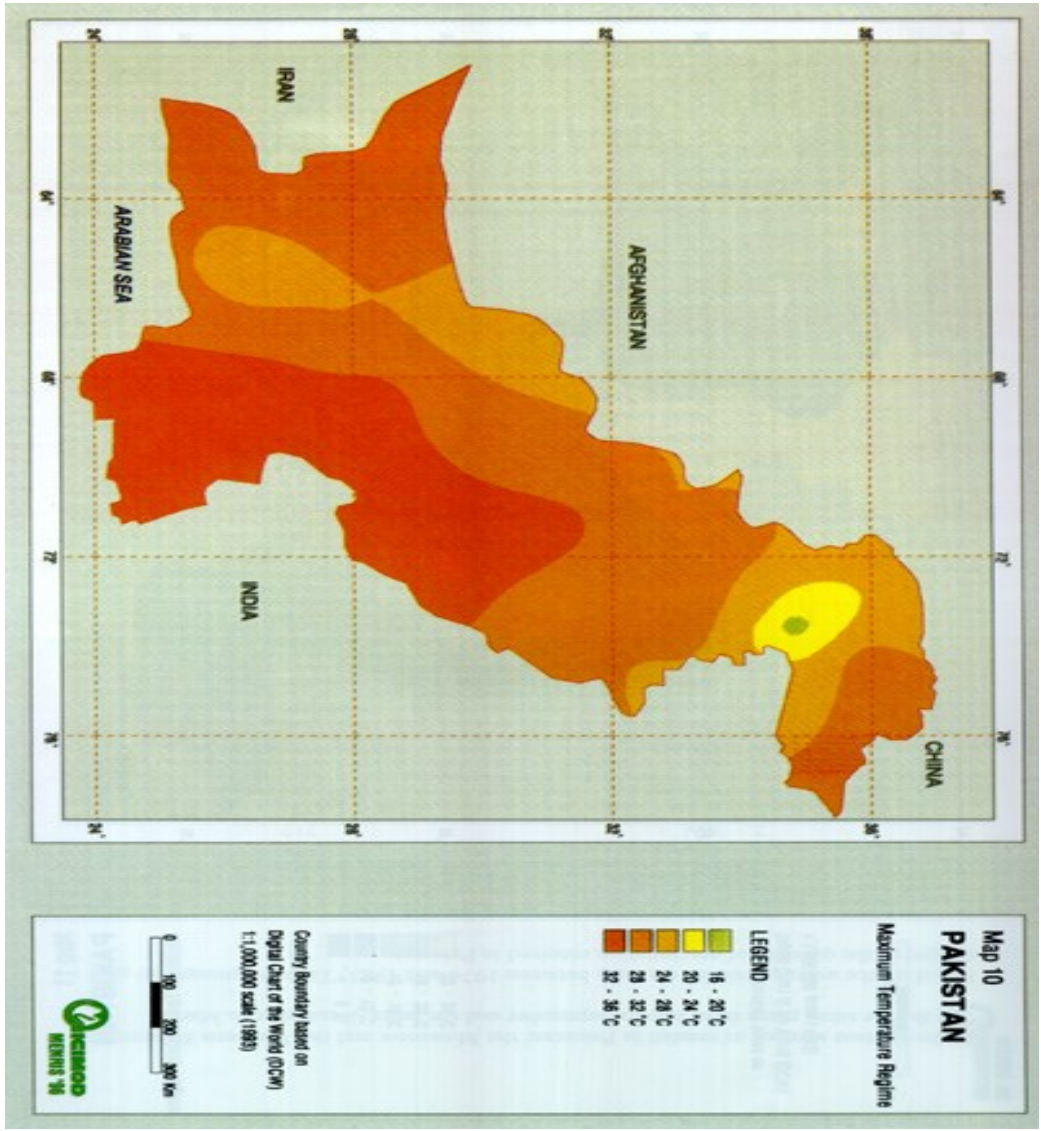

*Figure 1: Pakistan Temperature map showing that most of cultivated areas of central and lower Punjab and Sindh have very high mean maximum tempratures*

### NUMBER AND AREA OF FARMS BY SIZE OF FARM-2000

| Size of Farm (Hectares) | Farms Number | % | Farms Area Hectares | % | Avg. size of Farm Area (Hectares) |
|---|---|---|---|---|---|
| **Pakistan** | | | | | |
| All Farms | 6620224 | | 20437554 | | |
| Government Farms | 170 | | 30772 | | |
| Private Farms | 6620054 | 100 | 20406782 | 100 | 3.1 |
| Under 0.5 | 1290098 | 19 | 362544 | 2 | 0.3 |
| 0.5 to under 1.0 | 1099330 | 17 | 821245 | 4 | 0.7 |
| 1.0 to under 2.0 | 1425370 | 22 | 1981277 | 10 | 1.4 |
| 2.0 to under 3.0 | 966411 | 15 | 2256772 | 11 | 2.3 |
| 3.0 to under 5.0 | 890755 | 13 | 3442507 | 17 | 3.9 |
| 5.0 to under 10.0 | 580200 | 9 | 3891228 | 19 | 6.7 |
| 10.0 to under 20.0 | 260791 | 4 | 3324310 | 16 | 12.7 |
| 20.0 to under 40.0 | 77773 | 1 | 1955330 | 10 | 25.1 |
| 40.0 to under 60.0 | 15277 | * | 689070 | 3 | 45.1 |
| 60.0 and above | 14054 | * | 1682491 | 8 | 119.7 |

*Table 2: Farm holdings in Pakistan showing that majority of the land holdings in Pakistan are below the 12 acre mark [1]*

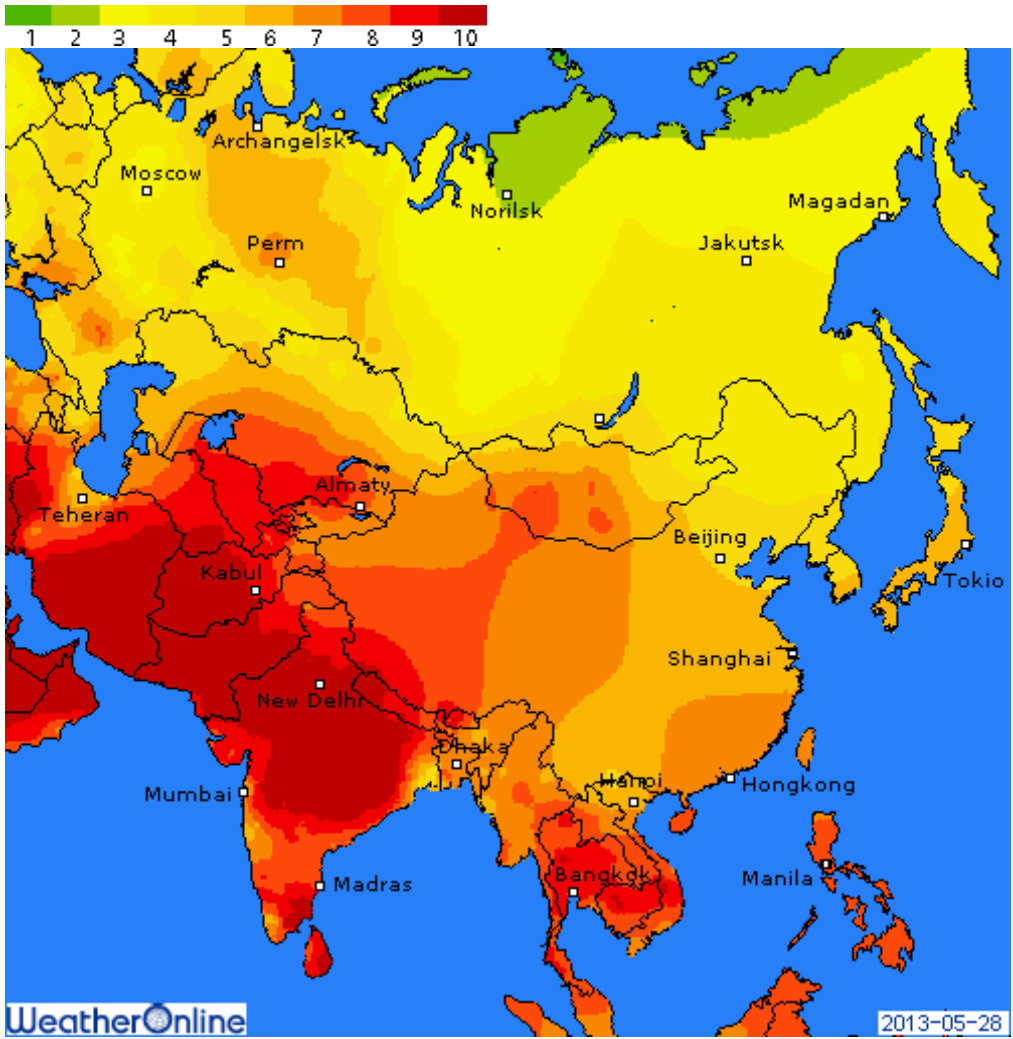

*Figure 2: UV index Pakistan showing the ammount of incident solar ultra voilet radiation [3]*

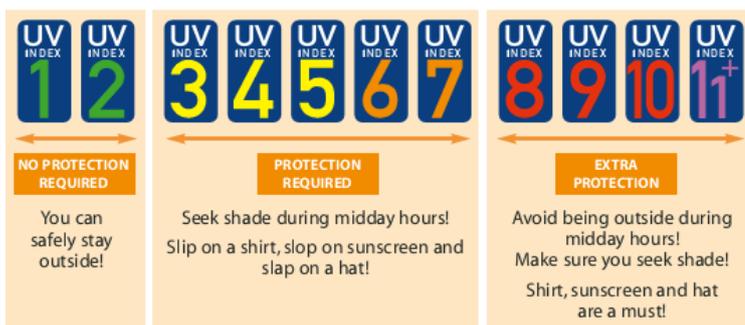

*Figure 3: Safety Instruction for different UV index values as given by the WHO [4]*

## 1.2 Proposed Solution
A number of solutions are viable for this problem farm mechanization, full scale farm automation and small scale farm automation.

Farm mechanization is a good solution to eliminate most of the labor but it is too costly as a combine harvester costs about Rs.4, 000, 000 and an average tractor at about Rs.6, 00, 000. 86% of the farmers are below 5 hectares (12 acres approx.) and they do not require a combine harvester as it is suitable for large holdings lying in a stretch and cannot afford a tractor(refer to Table 2: Farm holdings in Pakistan).

Full scale farm automation is a solution for large land holdings but is too costly for the small land holdings. Thus an optimum solution for the problem is an autonomous power reaper which is of small size and is thus cheap and suitable for the majority of the small farmers in Pakistan.

## 1.3 Orientation of the report

Chapter 2 will begin with introduction of Agricultural Robotics. There will be a literature review of the mechanical structure, vision and control algorithms.

In chapter 3 we will discuss the methodology in detail using block diagrams and flowcharts. The results of the tested and the proposed algorithms will also be displayed.

In chapter 4 we will discuss the results in detail and how they are of significance in our work.

In chapter 5 we will conclude our work and discuss some future perspectives.

In appendices we will provide some background information necessary regarding this project.

*Chapter 2*

# REVIEW OF LITERATURE

## 2.1 Agricultural Robotics

Agriculture is one of the most fundamental industries of the world as it is a source of food. Automation in this sector is led by the need to perform tasks with increased efficiency and accuracy. This need is also "technology-pushed" because the technology allows for the creation of fully unmanned farm vehicles.

The concept was originated in 1920's [5] with fully driverless systems using leader cables tested in 1950s and 1960s. In 1980s the vision systems were first used followed by GPS in 1990s [6]. In 1997 the concept of precision agriculture was introduced. Autonomous navigation systems are the heart of an agricultural robot.

| Institute (Country) | Sensor | Machine or test device | Performance results | Literature |
|---|---|---|---|---|
| University of Illinois, USA | Machine vision, GPS, GDS | Case 8920 MFD and 2WD Tractors | Vision guidance at 16 km/h on row crops | Zhang[9,10] Benson[11,12] |
| Stanford University, USA | GPS | John Deere 7800 Tractor | 1° accuracy in heading, line tracking accuracy with 2.5 cm deviation | O'Connor[13] |
| University of Florida, USA | GPS, laser radar | Tractor | Average error of 2.8 cm using machine vision guidance and average error of 2.5 cm using radar guidance | Subramanian[14] |
| University of Halmstad, Sweden | Machine vision, Mechanical sensor, GPS | Tractor with row cultivator | Standard deviation of position of 2.7 and 2.3 cm | Åstrand[15,16] |
| Bygholm Research Center, Denmark | Machine vision | Tractor | Accuracy of less than 12 mm | Søgaard[17] |
| University of Tokyo, Japan | FOG, Ultrasonic Doppler sensor | Tractor (Mitsubishi Co.) | Lateral displacement from the reference line was less than 10 cm at speeds of 0.7 to 1.8 m/s on a straight line | Imou[18] |
| National Agriculture Research Center, Japan | RTK GPS, FOG | PH-6, Iseki Co., Ehime transplanter | Less than 12 cm, yaw angle offset of about 5.5 cm at 2.52 km/h | Nagasaka[19] |
| BRAIN, Japan | Machine vision and laser range sensor | Tractor | Error about 5 cm at the speed of 0.4 m/s | Yukumoto[20] |
| Hokkaido University, Japan | GDS, laser scanner | Tractor | Average error less than 1 cm | Noguchi[21,22] |
| National Centre for Engineering in Agriculture, Australia | Machine vision | Tractor | Accuracy of 2 cm | Billingsley[23] |

*Table 3: Autonomous Navigation Systems developed in the recent times showing development of the idea [6]*

## 2.2 Vision Algorithms

We studied various algorithms to find out the suitable candidates for our task. We tested many but we will discuss only the most notable ones and explain as to why they were carried on or rejected.

### 2.2.1 Canny Edges

This edge detection operator was developed by John F. Canny in 1986. This is a multi-stage algorithm designed for realistic and effective edge detection reducing sensitivity to noise.

Explaining the algorithm briefly the first stage is a Gaussian filter which blurs out the noise in the raw data and thus is a crucial safeguard against noise. As, edges can be multi directional so canny operator uses four filters, horizontal, vertical and two diagonal to find all the edges with the help of following formulas.

$$G = \sqrt{G_x^2 + G_y^2}$$

Equation 1: Edge Gradient

$$\Theta = \arctan\left(\frac{G_y}{G_x}\right).$$

Equation 2: Edge Direction

The direction is assigned one of the four values 0°, 45°, 90°, and 135°. Based on this direction which check if the local maximum is attained in the region or not. For example:

- In case of 0° the region would lie in the north and south
- In case of 90° the region would lie on the east or west
- In case of 45° the region would lie on the north-west and south-east
- In case of 135° the region would lie on the north-east and south-west

After this stage we have thin edges which are further processed upon to attain the final image.

The major parameters of this operator are thresholds and the size of the Gaussian filter. Using a different combination of these we can attain a wide variety of results.

We can see from the results (Figure 4: Canny Edges result) that even though the canny operator works fine in some of the cases (Figure 5: Canny Edges result) but it is extremely susceptible to noise and thus cannot separate effectively in between crop residue (stalks that remain after the harvest), trees and the wheat crop. Another factor that slightly effects this method is that soil under the action of sun also cracks open due to presence of moisture, which are also detected in the algorithm.

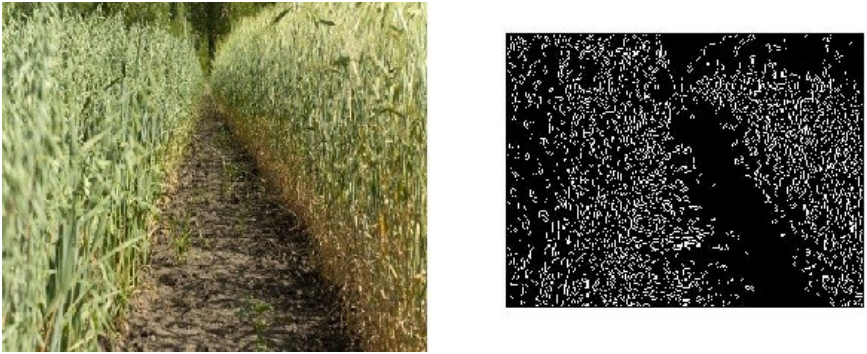

*Figure 4: Canny Edges result depicting satisfactory segmentation of ground from crop*

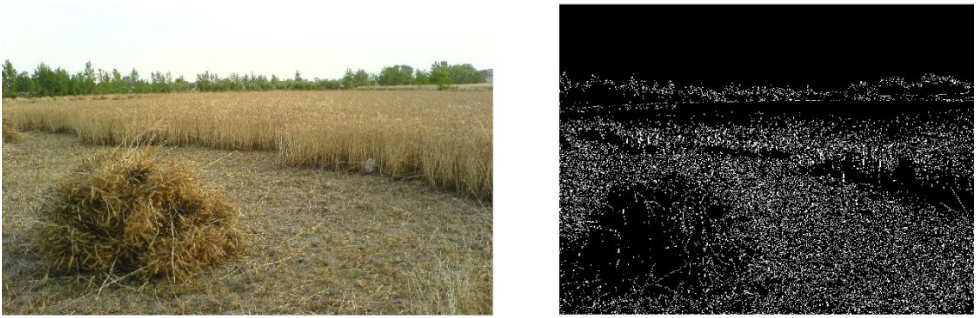

*Figure 5: Canny Edges result showing failure to segment out the residue from the crop*

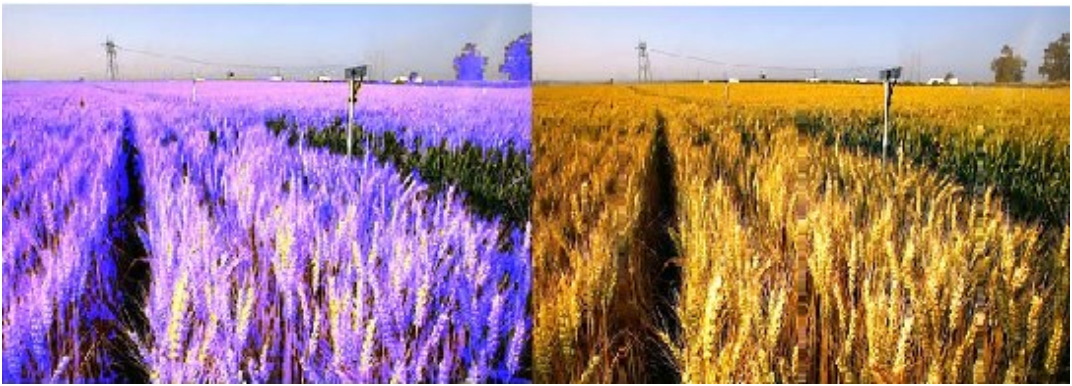

*Figure 6: Seed Template results showing the detection of wheat based on the seed template taken from same image*

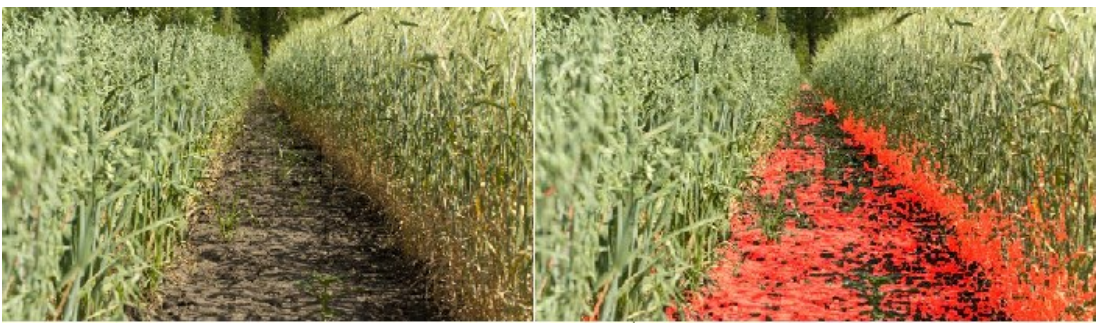

*Figure 7: Seed Template matching showing the detection of ground based on the seed template taken from same image*

### 2.2.2 Seed Template Matching

This method is based on one of the basic template matching techniques. We are taking a pre-defined template from the given image we know that has nothing except wheat. Then we use this block and match it with the whole of the picture. The matching has to be done on the basis of a property and many properties can be appropriately used but mean is of the best. We compare the mean of the template with a sliding window in the image and decide upon the similarity based on a threshold.

We also tested a slightly advanced variant of the above algorithm by introducing the slightly modified concept of Chi Squares as a measure of similarity or difference and then deciding on the basis of threshold.

$$\frac{(Mean_{window} - Mean_{Template})^2}{Mean_{Template}}$$

*Equation 3: Modified Chi-Squares*

Although this is quite a rough method but it is effective. It gives us very good segmentation results (Figure 6 and Figure 7) but the problems lie in the computational cost and the automation problems. Windowing operations generally take a lot of computational time and power both of which are a limited resource. One other problem is that for different images we require slightly different threshold values. The choice of the initial seed template also plays an important role in the final results. All of these problems were the reason that this method was discontinued despite its accurate results.

### 2.2.3 Gray Level Co-occurrence Matrix (GLCM)

We can clearly see that the major difference between the crop and the other irrelevant objects in the image is the texture. Crop has its own signature texture which is different from the tree, the sky the ground.

To exploit this difference we used the most basic texture analysis techniques, the Gray level co-occurrence matrix. A co-occurrence matrix is a table of information telling us how the pixel values were grouped with each other. As the grey scale pictures have 256 levels so such matrix for these pictures is 256 x 256.

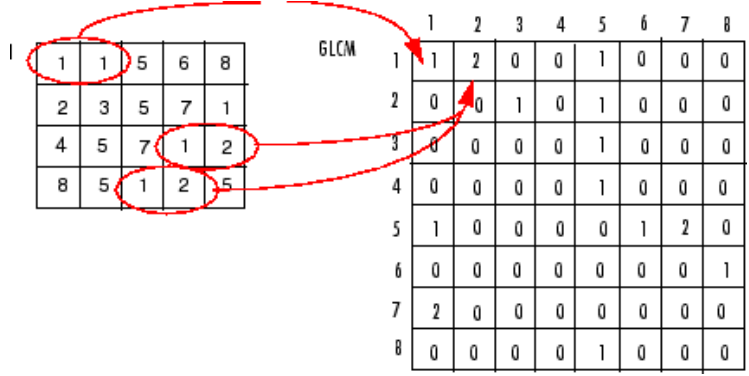

*Figure 8: Formation of a GLCM illustrating how we explore the meighborhood of number and record its co-occurance with other pixel values.*

The segmentation in this technique is carried out on the basis of different properties of the GLCM. The most fundamental of these properties are as follows:

$$\sum_{i,j} |i-j|^2 p(i,j)$$

*Equation 4: Contrast of GLCM*

$$\sum_{i,j} \frac{(i-\mu i)(j-\mu j)p(i,j)}{\sigma_i \sigma_j}$$

*Equation 5: Correlation of GLCM*

$$\sum_{i,j} p(i,j)^2$$

*Equation 6: Energy of GLCM*

$$\sum_{i,j} \frac{p(i,j)}{1+|i-j|}$$

*Equation 7: Homogeneity of GLCM*

This method also yielded satisfactory results but was dropped on the basis of its computation time and failure in figuring out a single property sufficient for all the cases. We can see from the results that different properties are required for segmentation in different cases.

### 2.2.4 Stereo Vision Techniques

The application of stereo vision techniques in agricultural robotics is more intuitive as we can think of it as a human being navigating the field. We now have the additional information of depth which can be extremely useful in estimating the distances in the field. These techniques have been used in assistive automation of farm machinery [7].

We dismissed these techniques because they require the use of a high quality stereo camera which is a costly equipment. Moreover to calculate the depth information in the real-time as we desire extensive computational power was required.

### 2.2.5 Color Space Transformations

The color space transformation is one of the most often used techniques in the object segmentation. The I1I2I3 color space was developed Ohta for the better segmentation between objects. This was improved by modifying it for plant-soil segmentation. [8]

$$\begin{pmatrix} i1_{new} \\ i2_{new} \\ i3_{new} \end{pmatrix} = \begin{pmatrix} 0.34 & 0.33 & 0.33 \\ 0.07 & 0.39 & -0.54 \\ -0.35 & 0.51 & -0.14 \end{pmatrix} * \begin{pmatrix} R \\ G \\ B \end{pmatrix}$$

*Equation 8: Modified I1I2I3 Color Space Transform*

The problem with this transform is that it is solely developed to differentiate between green plants and brown soil, but wheat when ripe turns golden-brown in color thus posing serious challenge to such algorithms.

## 2.3 Control Algorithms

As we are trying to control our robot on the basis of the sensor information acquired so the control system requires special attention. Controller is the heart of any autonomous system. Many control techniques have been implemented in the various farm automation systems over the year including neuro-fuzzy controller and other intelligent control techniques but we decided to use the simplest yet one of the most elegant techniques the PID controllers.

Let us shed some light on some types of controllers that can be used in the autonomous vehicles.

### 2.3.1 Neural Network Controllers

Neural networks or rather Artificial Neural networks were inspired by the studies into the working of the human cognition and thinking style. In contrast to the von Neumann model there is no separation between the memory and the data units but it works on the flow of signals through the basic units or the neurons and are thus similar to the biological neural systems.

The neural network has to be trained on training cases. A neural control process consists of two steps:

- System identification and modelling
- Controller design

A feed forward neural network with a nonlinear, continuous and differentiable activation function e.g. the sigmoid can be used to approximate any system while a recurrent network has countless applications in system identification.

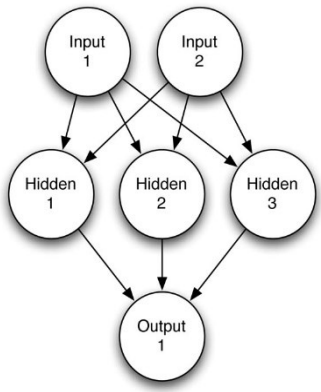

*Figure 9: Feed Forward Neural Network showing different layers of neurons the input, the hidden and the output layer*

### 2.3.2 Neuro-Fuzzy Controllers
The neuro-fuzzy logic systems are combination of the artificial neural networks and the fuzzy logic thus they combine the human reasoning with the learning ability and the highly interconnected structure of the neural networks. This functions on the principle that If-Then decision rules are extracted by the fuzzy reasoning and then the neural network is trained to work on the basis of these models which imparts two properties interpretability and accuracy.

### 2.3.3 PID Controller
The PID controller is a very elegant algorithm that is often used to make signals achieve certain set points. It is based on a very simple system design.

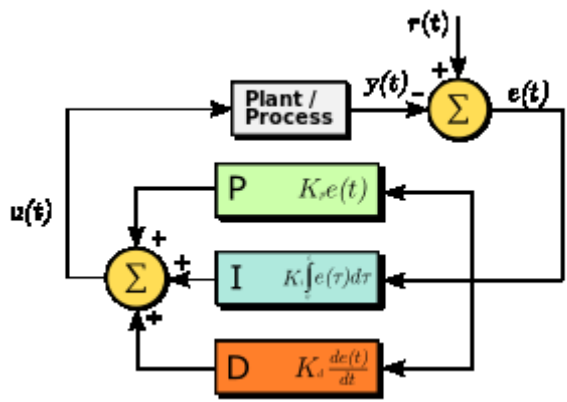

*Figure 10: PID system diagram showing how it interfaces with the plant system and showing its original components the proportional, derivative and integral*

$$u(t) = \mathrm{MV}(t) = K_p e(t) + K_i \int_0^t e(\tau)\,d\tau + K_d \frac{d}{dt} e(t)$$

*Equation 9: PID equation*

The proportional gain $K_p$ depends on the present error, Differential gain $K_d$ depends on the prediction of future errors based on the current rate of change, and the Integral gain $K_i$ on the values of the past errors. The values of these parameters need to be tuned by using various methods to attain the desired response.

## 2.4 Mechanical Considerations

Mechanical portion of our project was quite extensive so to cut down the work load we did not take extensive considerations into account in this project. There are many factors that will come into play when we go into an actual field. There are different kind of forces that will act on the tires due to the uneven terrain and the mud, which is quite different in its structural properties from concrete and other indoor materials.

The design of a vehicle used in farm is a separate subject and requires technical expertise and resources. Due to the lack of time, funding and most importantly the technical knowledge of farm vehicle designing we were unable to make our own platform. We had no choice but to go to market for available products that could satisfy our requirements.

We chose a battery powered toy car which had a payload of 40 kg and had two rear wheel drive motors. The battery time was 45 minutes. There were a lot of flaws in the design:

- Restricted angle of turning
- Plastic wheels
- Manual steering
- Wired logic electrical design

We decided to tackle these problems by using a motor to automate the steering, replace wired logic with programmable logic and manually modifying the design to extend the angle of turn but we could not modify it enough to successfully navigate an uneven terrain. We tested our algorithm on the basis of video shot in the real field to give a proof of concept.

We really hope that if the project is carried on a professional mechanical design is used to minimize power and cost, capable of navigating the field.

*Chapter 3*

# METHODOLOGY

In this chapter we will focus on the methodologies currently employed in the prototype of the autonomous power reaper. We will discuss the vision and the control algorithms separately at first and then we will shed some light on the integration of both into a working prototype.

## 3.1 Vision Algorithm

The vision algorithm that we have used is a result of extensive experimentation and testing with various classical techniques on the images of wheat farms. We combined some of the techniques discussed in the previous chapter to come up with a new algorithm which has been exceedingly successful.

Now we will discuss the algorithm in detail starting with the idea and proceeding to the implementation.

### 3.1.1 The idea

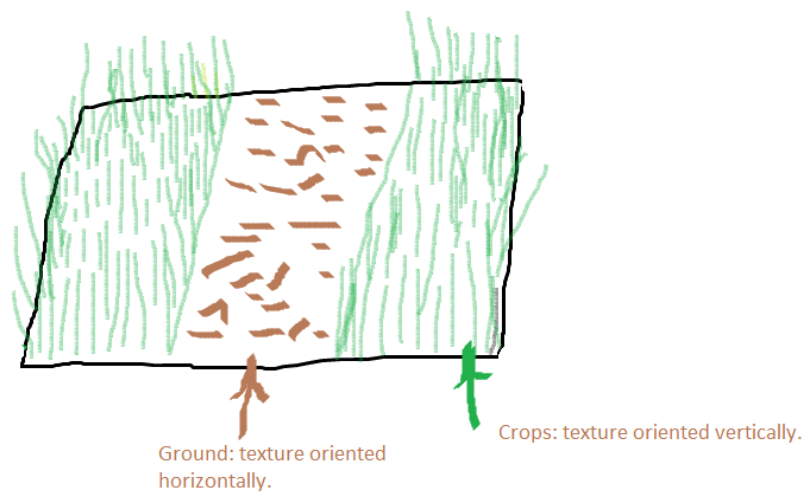

*Figure 11: Texture orientation in Field showing the vertical texture orientation in crops and the horizontal texture orientation on the ground*

The figure (Figure 1) above explains a key point, the texture of the ground is horizontally oriented while that of the crop is vertically oriented. This point also gives us an advantage as the crop that is cut is also laid on the field in the horizontal direction. So, if this point is exploited then we will be able to segment out both the ground and the harvested crop.

Another problem that we face is that the land and the ripe wheat crop have minute difference of color in the sunlight. To solve this problem we decided to use color space transformation and spread the contrast among different shades of yellow for better segmentation. Now let us discuss the algorithm in detail.

### 3.1.2 Detailed description

The vision algorithm is a three stage algorithm. The system diagram below gives a better insight into the technique.

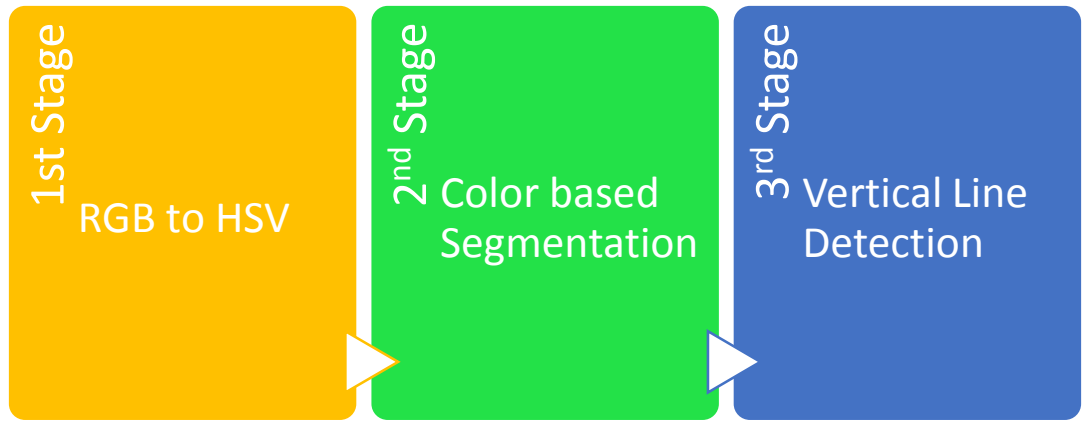

*Figure 12: Flow chart of vision Algorithm showing different stages of processing*

### 3.1.2.1 RGB to HSV

HSV is a more intuitive color space for us to understand than the RGB. The HSV separates the chromaticity (Hue) and the luminosity (Value) rather than mixing them up with the color values. HSV is more useful as we desire to achieve a color based segmentation in in the HSV the color is represented in the H or the hue channel as a 360° circle.

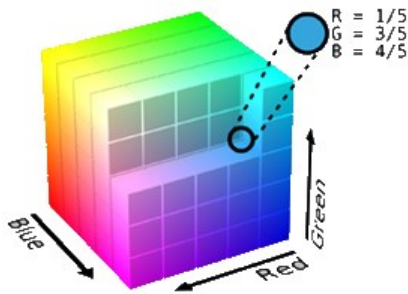

*Figure 13: RGB cube showing the positions of different colors. What this picture fails to show is that shades of a color can be separated oddly in the cube making segmentation difficult*

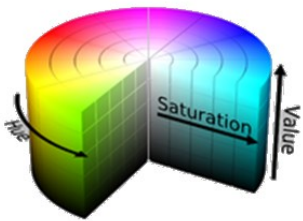

*Figure 14: HSV Cylinder showing how the color distribution is easy to manipulate and intutive*

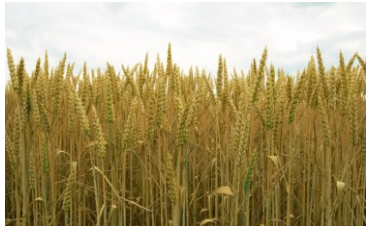

*Figure 15: Original RGB image*

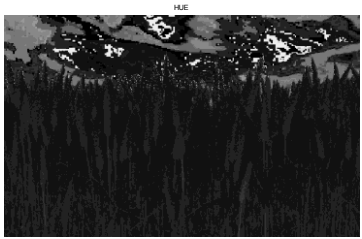

*Figure 16: HUE channel showing which portions have same color*

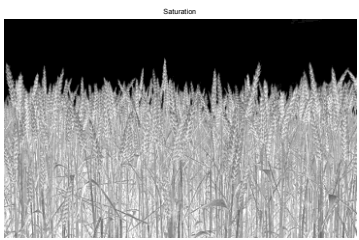

*Figure 17: Saturation showing which region consist of what purity of primary colors*

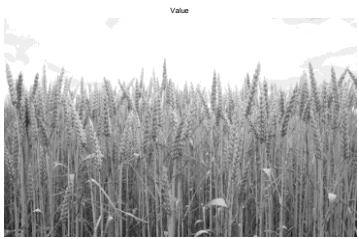

*Figure 18: Value shows the effect of luminosity that has affected the image. This is not accurate but is an educated guess based on the light source models*

In the HSV cylinder we now need to separate out the yellow color thus removing all the unnecessary background information only keeping the wheat and the ground for further processing. After seeing the three channels, the hue (Figure 16), saturation (Figure 17), value (Figure 18), of the original image (Figure 15), a question can arise as to why do not use the saturation channel rather than the hue channel.

The answer to this question lies in the definitions of the hue and the saturation. Hue is defined as the amount by which the stimulus varies from the pure primary colors (red, blue, green) while the saturation is defined as the intensity of the color and its distribution across the spectrum. Thus a single primary color of a single wavelength of very high intensity will have a high saturation.

In saturation two objects of different colors can theoretically have same values which is not possible in the hue domain unless they have exactly the same color. This is the reason due to which we have selected the hue which is intuitively more suited for our application.

### 3.1.2.2 Color Segmentation

We have decided upon the preprocessing techniques, now we have the task of segmenting out the wheat and the ground on the basis of the hue values. As we have seen in the previous figure of the HSV cylinder (Figure 14) the region of interest will lie in some portion of this color space. Intuition and experimentation has shown that the region of interest lies in an area of the cylinder similar to the one shown in the figure below.

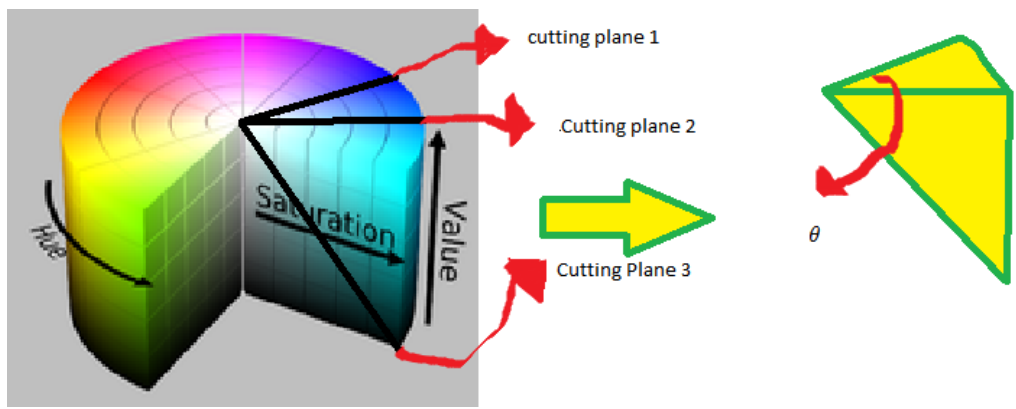

*Figure 19: Segmentation of HSV cylinder by three cutting planes two perpendicular to hue and one in the saturation and value plane*

As we can see that the hue is circular scale, so to segment out the desired values we require two planes in the perpendicular to the hue axis. Both of these planes (cutting plane 1 and cutting plane 2) will be at a certain angle (θ). The value of this angle (θ) determines the range of colors which will be segmented out during this phase of segmentation. The equations of the cutting planes can be denoted as follows:

$$cutting\ plane\ 1: Hue = \varphi_1$$

*Equation 10: Cutting Plane 1*

$$cutting\ plane\ 2: Hue = \varphi_2$$

*Equation 11: Cutting plane 2*

$$\theta = \varphi_2 - \varphi_1$$

*Equation 12: Range of colors segmented*

For the best results segmentation must also be carried out in the saturation and value domains because it is not only the hue that makes up a color but the mixture of all three properties which define the color the sensors read. So for the segmentation in the saturation and the Value axis we define a plane in the Saturation and the Value axis. The cutting plane 3 plays an important role in the segmentation as if we can fine tune the fitting of this plane we can remove lot of unwanted noise in this step and thus we can have better results in the nest domain. The constants, '**a**' and '**b**' are variable and have to be tuned at the initialization. The equation of this cutting plane is as follows. We should keep in mind that in the HSV domain:

$$Hue \in [0°, 360°]; Saturation \in [0,1]; Value \in [0,1]$$

*Equation 13: Range of Axis in HSV*

$$a * Saturation + Value = b$$

*Equation 14: Cutting Plane 3*

After we have successfully segmented out the crop and the ground from the rest of the background (Figure 23), comes the crucial step of segmenting out the crop from among the resulting image. This is done on the basis of simple thresholding (Figure 23).The threshold is manually decided in this algorithm. This gives us all the ripe wheat crop present in the image.

### 3.1.2.3 Vertical Line Detection

After the step that we have concluded above, we have a picture with all the wheat pixels marked white but this information alone is not enough. As we can see in the image (**Error! Reference source not found.**) that the harvested crop is present in the field along with the ripe crop that has not yet been harvested.

This is an evident problem as it will also be detected as wheat pixels. This must not be detected as it has already been reaped, so we have completed our work in this area. Moreover we do not want to pass over the harvested crop as it will be destroyed. So, we need to separate out only the crop that is still standing rather than the crop that has been harvested.

The solution to this problem is to detect vertical lines as the harvested crop lies horizontally. We can do this by several ways including Hough transforms and morphological operators. We selected the morphological operations for fast computation.

The morphological operation that we select is erosion. The choice of the structuring element is also very crucial for good results. Now let us discuss this matter in some detail. As we know that we are detecting vertical lines so the choice of structuring element would evidently be the straight line.

Although the choice looks quite straight forward but there some other factors which we must take in to considerations. Take a look at **Error! Reference source not found.** , we can see that wheat plant is not perfectly vertical but there is a slight tolerance in the angles. So this causes a problem as we have to cater for a range of angles which calls for structuring elements of different tilt as in **Error! Reference source not found.**. Another potential problem is that as in **Error! Reference source not found.** we can see that the residue, the portion of wheat stalk left by the reaper during the harvest will also be detected as vertical lines. For this purpose we have placed a check on the height of the straight line which is variable for different fields as the different varieties of wheat have different heights.

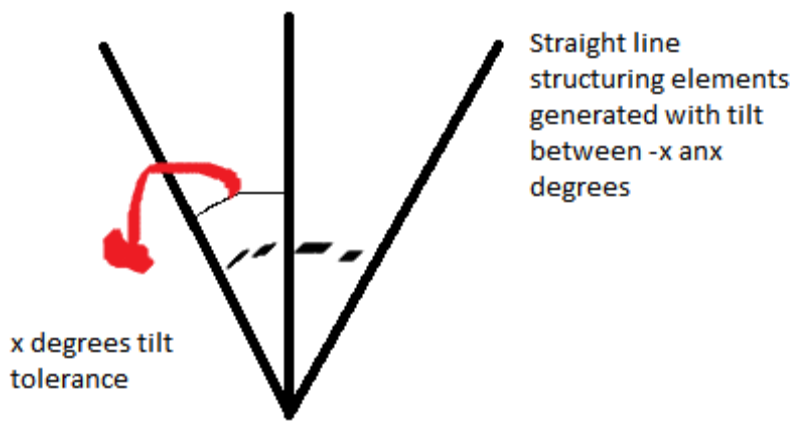

*Figure 20: The structuring elements generated for erosion. The tilt factor of x degrees is variable.*

## 3.2 Control Algorithm

Control is very important as our system has to be controlled in an uneven terrain. We have three actuators in our system which are the two back wheel motors and the steering motor which is responsible for the heading.

We have no feedback on the rear wheel drive motors as we are solely depending on the vision system, but the heading is important in our application so we must have a foolproof system of feedback for this. Out of the two possibilities, encoded feedback from the motor and the vision based heading information assisted by visual markers, we choose the encoder method as it is easy to implement.

We make a simple resistive encoder from a potentiometer and mount it on the steering rod. The rotation causes the potentiometer to rotate. We have mounted it in the form of a resistive divider to our controller. Thus we read the voltage at the center tap after applying source and ground to the other two. This gives us an idea of the position of the steering. We have divided the whole steering angle in to six equal steps three on the right and three on the left. Before the start of harvesting the code must be tuned according to the values of the potentiometer at the extremes.

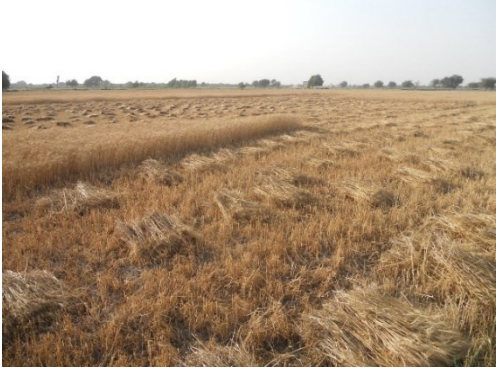

*Figure 21: A wheat field during harvest showing standing crop, residue and harvested crop lying horizontally*

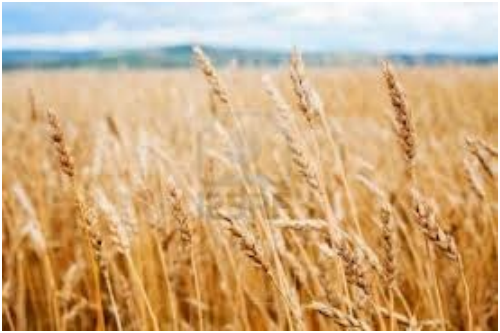

*Figure 22: A picture showing that the wheat stalks slightly tilted from the vertical*

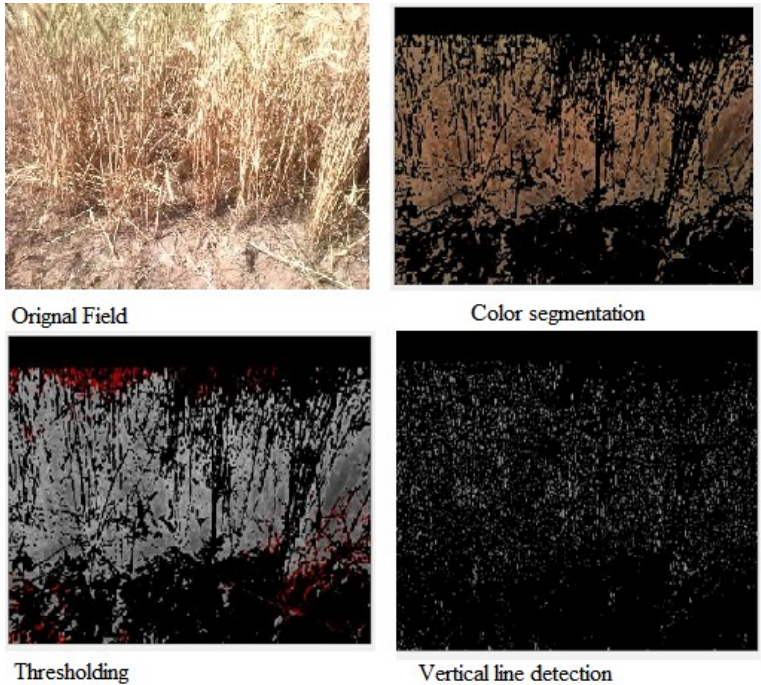

*Figure 23: The results of the different stages of the proposed algorithm showing the working in detail*

We have implemented a PWM based speed control in the rear motors. We have a PID controller implemented which takes the set points provided by the code and vary the values of the rear motor PWM. The front motor position are hard coded based on the resistive encoder values. A problem arose due to the amount of load on the front wheels that we could not implement the PID control based heading control. This was due to the fact that at lower values of PWM the steering motor did not work.

## 3.3 Integration and Assisted GPS

There are many styles which are used to cover the area of the field during the harvest but we have to select the one which is least heavily taxing on the controllers and the hardware. So the method that we have selected is depicted as follows in Figure 24.

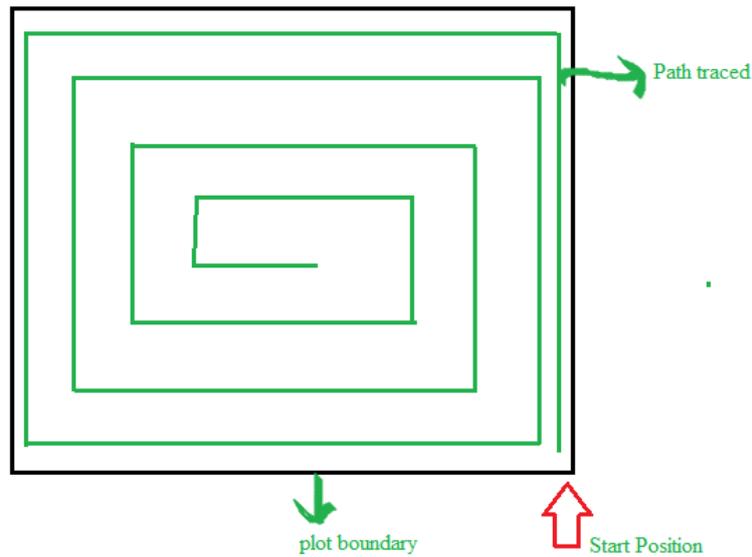

*Figure 24: this figure shows the method that we have adopted to cover the area of the whole field.*

Integration of the vision and the control algorithms was done using some basic pin hole camera physics concepts. Now we will discuss several maneuvers that are used in the field and how they are performed as a result of the visual-motor coordination.

To keep the matters simple we start from a supposition that we are a corner of the field alongside one border and our field of view comprises of wheat in one half and the ground in the other. One of the most important maneuvers in the field is to move towards the crop and to maneuver the robot as to keep it in the region where the crop has not yet been harvested. We use a simple concept for this action. We suppose the area where the wheat is detected to be a 2-D body. We find a centroid of that body. The cameras field of view has been hypothetically divided into three regions as shown in Figure 25.

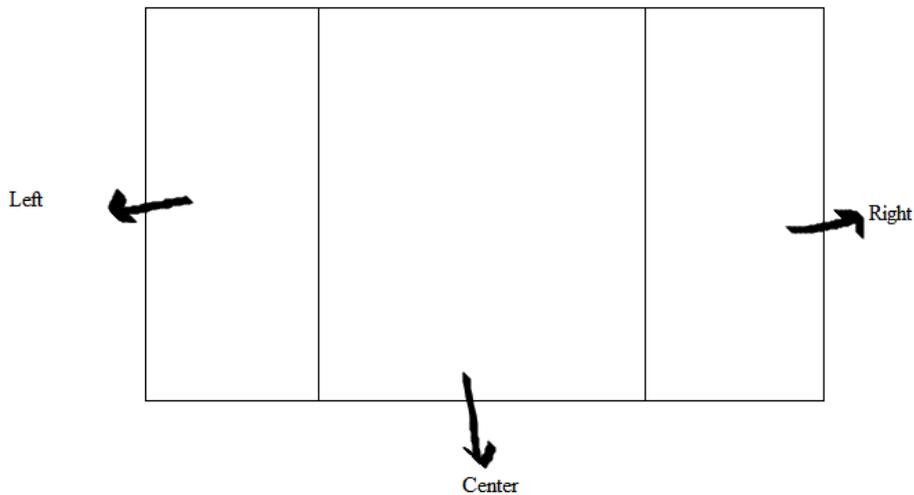

*Figure 25: This diagram shows hypothetical segmentation of field of view of the camera into three regions*

The decision is made according to the location of the centroid in the field of view. This process done on every frame repeatedly ensures that the robot would move such as to keep the region of the wheat in the center portion of its field of view.

Another important action that is required in the field is the detection of the separation between two separate plots of wheat. For this we require the basic knowledge of the pinhole camera model. From the pinhole camera model we know that the object which is farther in the actual world appears shorter than the object that is nearer to the camera (Figure 26). Using this fact we can determine when we have reached the separation between two plots.

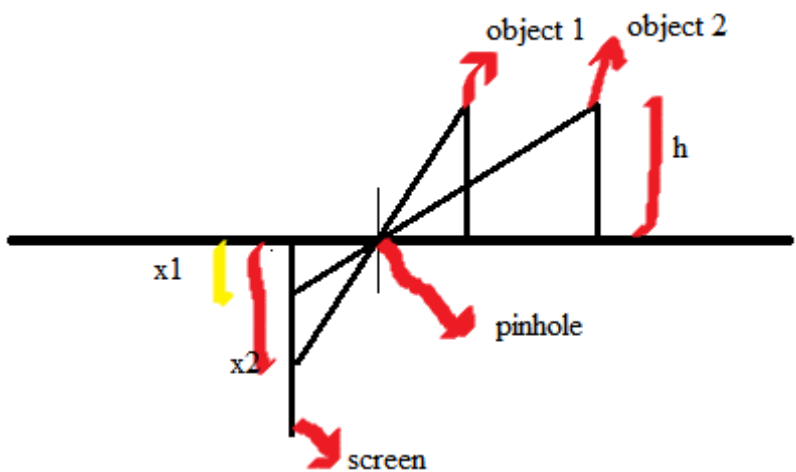

*Figure 26: This diagram shows the incident rays of the two objects of same height and there projections proving that the near objects are higher in the image*

As we are cutting in the plot 1 wheat plants have negligible separation thus we have all the wheat plants terminating at a constant height in the image, but as we reach the separation and cut the last of the wheat the next wheat plant will be after a considerable gap so the wheat belonging to the plot 2 will appear at a smaller height in the image. This sudden change of the crop roof can be used to detect the separation and take a turn so as to remain in the plot.

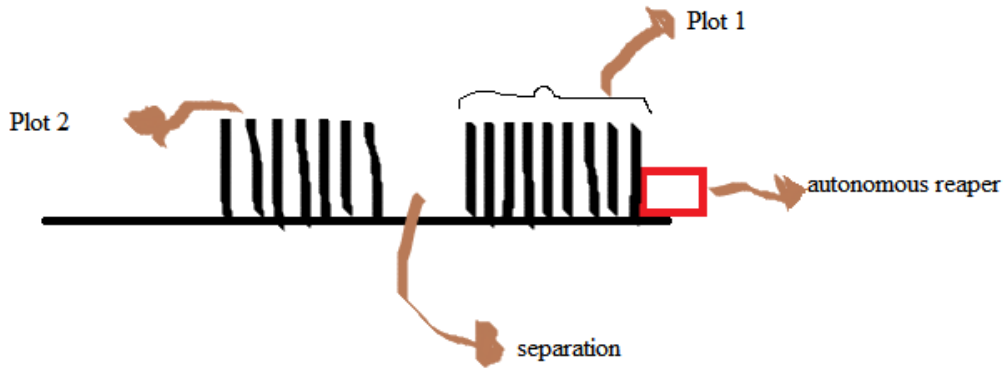

Figure 27: This figure helps in visualizing how the robot will approach the separation head-on while harvesting

As we are processing frames of videos to extract the information so there is time interval between the arrivals of data. So there is a margin of error that we may not detect the boundary. The boundary detection is important because we do not want cross into neighboring fields which may belong to other farmers and cause unwanted damage. Moreover the separations between fields are often raised or protected with shrubs. The height of separation may not be enough to be detected in the camera, because we have set the angle of elevation according to wheat plant. These can cause damage to the vehicle which is extremely undesired.

To cater for the boundaries we have the support of Assisted-GPS which comes in mobile phones today. We have used Assisted-GPS as it gives us two very key advantages. There are two types of assistance:

- Mobile Station Based
- Mobile Station Assisted

The mobile station based assistance is that we can download the orbital and other necessary information from the AGPS server installed by the network operator rather than at the 50 bits/sec link provided by the satellites. Moreover we can connect to satellites even in poor conditions based on the server data. [9]

The mobile station assisted feature is that based on the GSM tower co-ordinates and the information of the ionosphere conditions as collected by the towers , we can calculate our position very accurately as compared to stand alone GPS. [9]

We first trace the boundary of the plot by the receiver to store the co-ordinates of the boundary and then use them as a check to prevent the robot from crossing them under any condition. We can also save memory by an assumption that the field has straight boundaries and then storing only the boundary points. We can generate rest of the points based on the equations of straight lines.

*Chapter 4*

# RESULTS AND DISCUSSION

In this chapter we will discuss results of the algorithm that we have developed in the previous chapter. We will discuss the two most common cases that will occur in the field and we will try to explain the importance of each case with comprehensive detail

## 4.1 Case 1: In the middle of the field with no gaps and separation

Although this case is difficult to occur but we will discuss it as it is helpful in demonstrating the working of the algorithm.

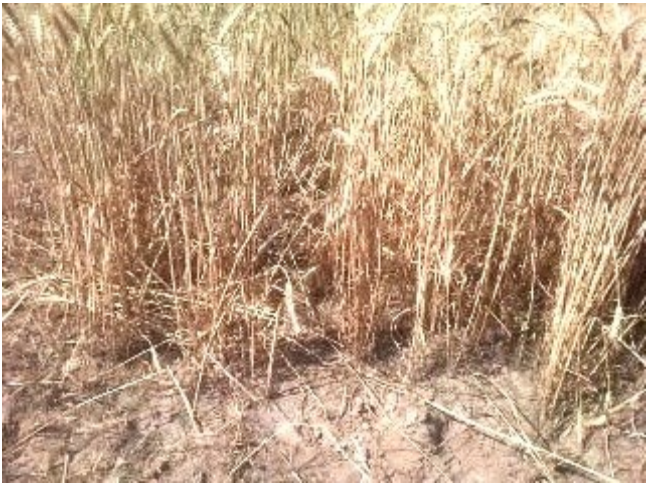

*Figure 28: Original image from the frame captured in the field at 2:38 PM in sunlight. It has some portion of ground and a major portion of Wheat.*

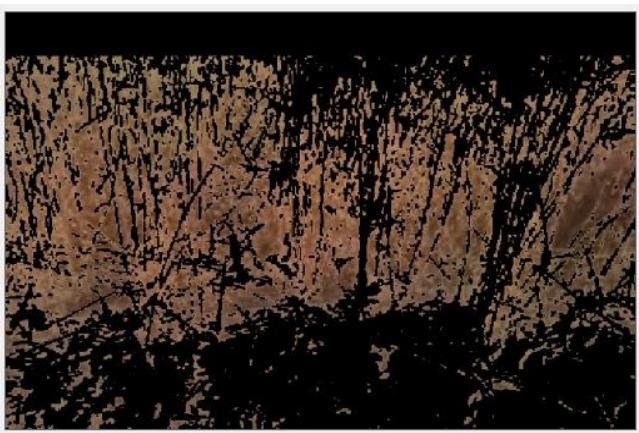

*Figure 29: This image shows the result of color segmentation in hue channel. There is a problem as some of the portion has not been detected although it is a part of the wheat. This is due to the directly incident sunshine which turns it into a shade of white*

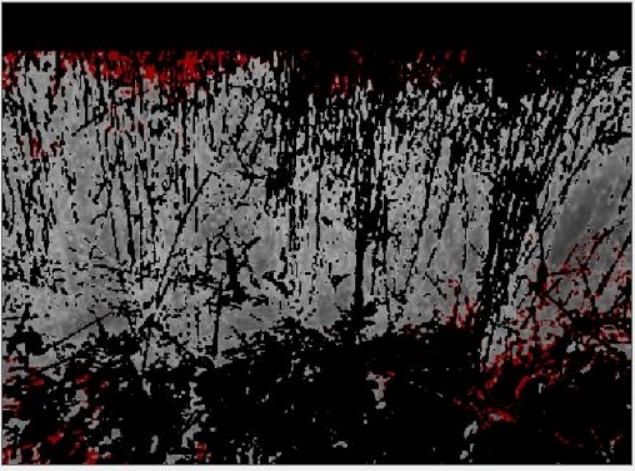

*Figure 30: This image shows the result of thresholding on the previous image. We can see that it has segmented out wheat as white.*

The anomalies can be removed by assuming that the ground pixels always be at the bottom half of the image with surrounding cluster of red pixels.

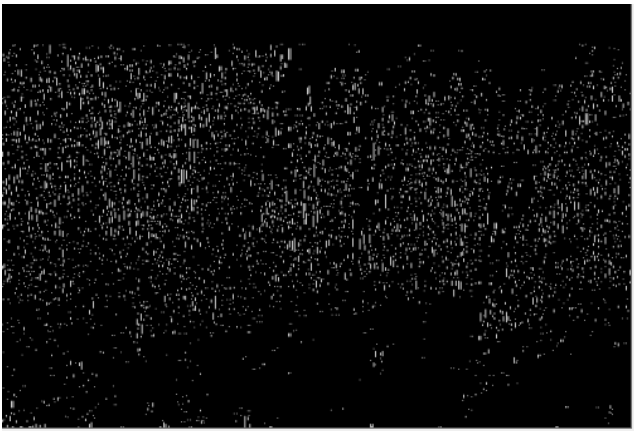

*Figure 31: This image shows the result after vertical line detection. We can see a lot of small lines but the feature to notice is that the concentration of the lines is the greatest in the region where there is crop standing*

## 4.2 Case 2: Crop on both sides of a separation and a tree

This case emulates one of the most difficult farm conditions. It is dusty, sunny and there is separation between two plots and there is a tree in the image. The results of this image will truly demonstrate the abilities of our algorithm in real field like conditions.

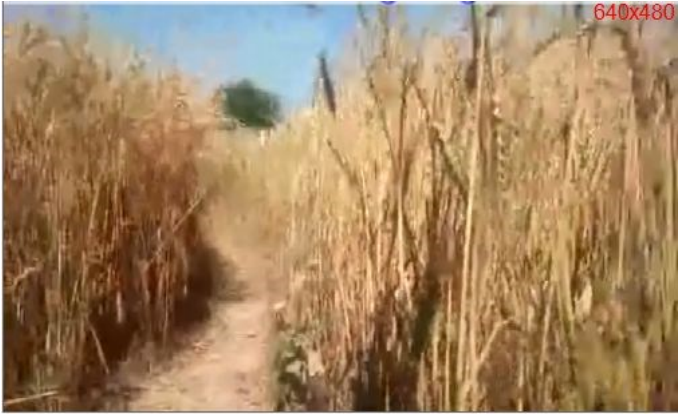

*Figure 32: This image is the original frame capture from a field at 3:00 PM under bright sunny conditions and dusty environment. This image has a separation between the two plots and a tree in the middle.*

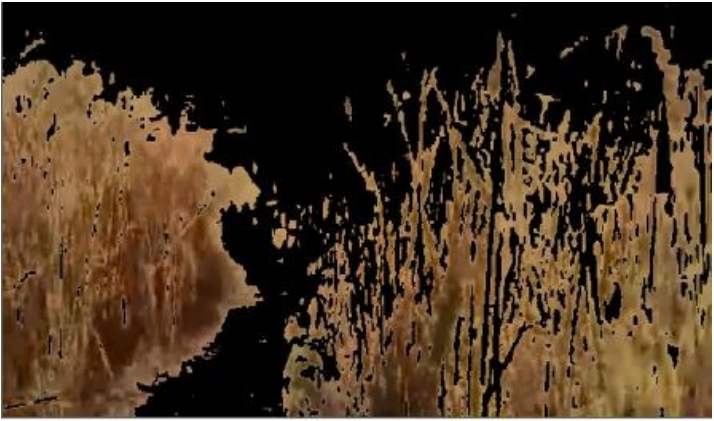

*Figure 33: This image is the result of color segmentation applied on the previous image in the hue channel. The tree and the sky and all other from the background and have been segmented out even some portion of ground to directly incident sunlight*

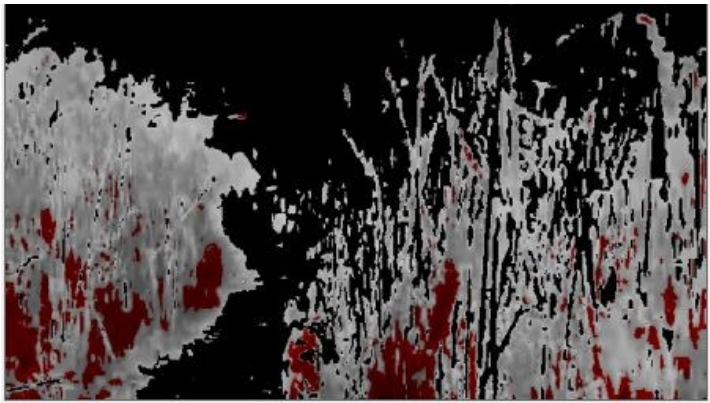

*Figure 34: This figure is the result of thresholding performed on the previous image. The thresholding causes the lower portions of wheat to be classified as ground. This is caused due the dust in the environment.*

To solve the problem we see in Figure 34 we can assume that if ground is detected enclosed or bounded by wheat we must also go there as it most probably is wheat and the results can be a result of some malfunction.

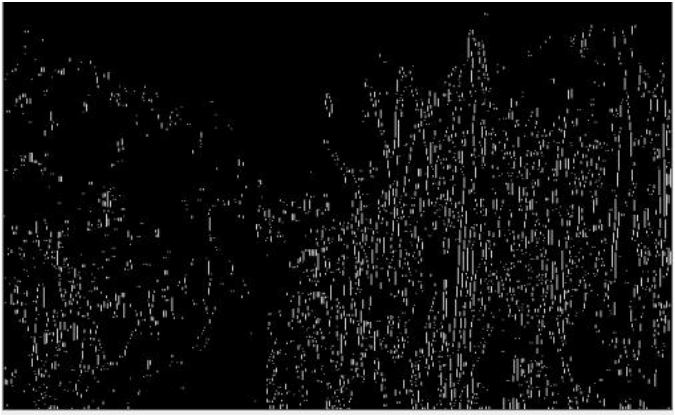

*Figure 35: This figure is the result of vertical line detection performed on the original picture we can see that the concentration of vertical lines is high in the region wheat and very low in the region of ground*

# CONCLUSION AND FUTURE PROSPECTS

We had started this project with an idea that we will introduce this field in Pakistan. Pakistan is primarily an agricultural country and its prosperity is tied closely to this sector. As in the earlier days the mechanization brought a revolution in the fields we believe that Agricultural Robotics or rather 'Agribotics' has the potential to revolutionize the agricultural sector not only in the wheat farming but also in other crops, horticulture, livestock and forestry. So, research and development in this field must be carried out in this field to secure the future of our nation.

## 5.1 Conclusion

We have been successful in developing an algorithm which can segment out ripe wheat crop from the ground, residue and harvested crop lying in the field. We have used this algorithm to run a prototype vehicle on the basis of video recordings of the actual field.

We have failed to develop a mechanical body that is capable of movement in the field conditions due to lack of funding and technical expertise. We also have several sensory issues which can be solved by using better quality sensors.

We could not achieve the fully autonomous robot due to variability of light and weather conditions. We have to manually tune the parameters of the algorithm. Despite this we have reduced the workload by a significant amount.

## 5.2 Future Prospects

This project has many future prospects but I would only discuss a few important ones which are as follows:

- Extension to other crops: This project can be easily extended to other similar crops like rice. To extend to sugarcane, and millets several mechanical considerations will have to be made
- Auto-tuning of algorithm parameters: This can be implemented based on the light conditions to make it fully autonomous
- The idea can be extended to swarm robotics by making smaller robots to cooperate and work efficiently and cost effectively in large farms.

# CODE IMPLEMNTATION OF METHODOLOGIES

## A1.1 MATLAB code

```matlab
close
clear
clc
%loading image
img = imread('results.jpg');
% defining the min length for vertical line detection
x = 10;
f = 3;
% RGB to HSV Conversion
imh = rgb2hsv(img);

[row col ch] = size(imh);
% color based segmentation in the hue domain
for r = 1:row
    for c = 1:col
        if(imh(r,c,1)>0.18 || imh(r,c,2)<(0.1*imh(r,c,3)))
            imh(r,c,3) = 0;
            imh(r,c,1) = 0;
        else
            if (imh(r,c,1)<(0.9 * imh(r,c,2)))
                imh(r,c,2) = 0;
            else

                imh(r,c,1) = 0;
                imh(r,c,2) = 1;
            end
        end
    end
end

imb = hsv2rgb(imh);
imshow(imb)
% threshholding to separate wheat and ground
I = im2bw(imb,0.45);
% generating vertical line structs with tilt
struct1 = strel('line',x,90);
struct2 = strel('line',x,95);
struct3 = strel('line',x,100);
struct4 = strel('line',x,85);
struct5 = strel('line',x,80);
im1 = imerode(I,struct1);
```

```matlab
% Erroding for vertical line detection
im2 = imerode(I,struct2);
im3 = imerode(I,struct3);
im4 = imerode(I,struct4);
im5 = imerode(I,struct3);
imf = im1 + im2 + im3 + im4 + im5;
subplot(1,2,1),imshow(img);
subplot(1,2,2),imshow(imf );
```

## A1.2 C# CODE

```csharp
private void ProcessFrame(object sender, EventArgs arg)
{
    // capturing video
    Image<Bgr, Byte> ImageFrame = capture.QueryFrame();
    // Measuring frame size
    width = ImageFrame.Width;
    height = ImageFrame.Height;

    res.Text = width.ToString() + "x" + height.ToString();

    Image<Bgr, Byte> img = ImageFrame;

    //Convert the image to grayscale and filter out the noise
    Image<Gray, Byte> gray = img.Convert<Gray,   Byte>().PyrDown().PyrUp();
    // Applying Canny edges
    Gray cannyThreshold = new Gray(30);
    Gray cannyThresholdLinking = new Gray(50);
    Gray circleAccumulatorThreshold = new Gray(100);

    Image<Gray, Byte> cannyEdges = gray.Canny(cannyThreshold, cannyThresholdLinking);

    // Converting the image to HSV scale
    Image<Hsv, byte> imhsv = new Image<Hsv, byte>(ImageFrame.Bitmap);
    Image<Gray, byte> imgr1 = new Image<Gray, byte>(imhsv.Bitmap);
    Image<Gray, byte> imgr = new Image<Gray, byte>(imhsv.Bitmap);
```

```csharp
// Color based Segmentation in HSV domain
for (int i = 0; i < height; i++)
{
 for (int j = 0; j < width; j++)
 {
  imgr.Data[i, j, 0] = 255;
 }
  }

for (int i = 0; i < height; i++)
{
 for (int j = 0; j < width; j++)
 {
  if ((imhsv.Data[i, j, 0] > c1  && imhsv.Data[i, j, 0] > (c1 - 1) ) || (imhsv.Data[i, j, 1] < (c2 * imhsv.Data[i, j, 2]) ))
  {
    imhsv.Data[i, j, 2] = 0;
    imhsv.Data[i, j, 1] = 255;
    imhsv.Data[i, j, 0] = 0;
   }
else
    if (imhsv.Data[i, j, 0] < (c3 * imhsv.Data[i, j, 2]))

     {
        imhsv.Data[i, j, 1] = 0;
        imgr.Data[i, j,0] = 0;
     }
     else
     {
        imhsv.Data[i, j, 0] = 0;
        imhsv.Data[i, j, 1] = 255;
        imgr.Data[i, j, 0] = 255;
     }

  }
 }
// Thresholdin the image to segment
imgr = imgr.ThresholdBinary(new Gray(63), new Gray(255));

imgr = imgr.Canny(new Gray(3), new Gray(50));   //try 180 - 120

imgr1 = imgr1.ThresholdBinaryInv(new Gray(80), new Gray(255));

//Creating Straight line Structuring element
StructuringElementEx struc1 = new StructuringElementEx(1, stel, 0, 0, Emgu.CV.CvEnum.CV_ELEMENT_SHAPE.CV_SHAPE_RECT);

Image<Gray, byte> cannyedge2 = cannyEdges;
```

```csharp
// Creating structuring elements with tilt
byte[, ,] arr = new byte[5, 2, 1];
            byte x = 3;
            int y = x;

            Image<Gray, byte> img = new Image<Gray, byte>(@"D:\pic.jpg");
            Image<Gray, byte> img1 = new Image<Gray, byte>(arr);
            imageBox1.Image = img;
            img1 = img1.Rotate(30, new Gray(255),false);
            imageBox2.Image = img1;
            List<int>  asdf = new List<int>();

// Erroding the image with the struct for vertical edge detection
CvInvoke.cvErode(imgr, cannyedge2, struc1, 1);

CamImageBox.Image = ImageFrame;
cambox2.Image = cannyedge2;
imageBox1.Image = imhsv;

}
```

# References

[1] Ministry of Finnance, "Economic Survey of Pakistan 2011-12," Government of Pakistan, Islamabad, 2012.

[2] K.A. Quraishi, Buland Akhtar, Ch Muhammad Aslam, Malik Ikram and Asghar Ali, "SOCIOECONOMIC EFFECT OF INDUSTRIALIZATION ON THE SURROUNDING RURAL AREAS WITH SPECIAL REFERENCE TO AGRICULTURE: A CASE STUDY OF ISLAMABAD DISTRICT," *Pakistan Journal Of Agricultural Sciences,* vol. 31, no. 3, pp. 236-240, 1994.

[3] WeatherOnline Ltd., "Pakistan UV index:WeatherOnline," WeatherOnline Ltd., [Online]. Available: www.weatheronline.co.uk. [Accessed 28 May 2013].

[4] World Health Organization, "Global Solar UV Index- A Practical Guide," WHO, Geneva, 2002.

[5] Ming Li, Kenji Imou, Katsuhiro Wakabayashi, Shinya Yokoyama, "Review of research on agricultural vehicle autonomous guidance," *International Journal of Agricultural and Biological Engineering,* vol. 2, no. 3, pp. 1-16, 2009.

[6] K. Siciliano, "Robotics in Agriculture and forestry," in *Handbook of Robotics*, Berlin, Springer, 2008, pp. 1065-1077.

[7] F. Rovira-Más, S. Han, J. Wei, J. F. Reid, "Autonomous Guidance of a Corn Harvester using," *Agricultural Engineering International: the CIGR Ejournal,* vol. 9, no. 7, pp. 1-13, 2007.

[8] Isabelle Philipp, Thomas Rath, "Improving plant discrimination in image processing by use of different color space transformations," *Computers and Electronics in Agriculture,* vol. 35, pp. 1-15, 2002.

[9] Wikipedia, "Assisted GPS-Wikipedia," Wikipedia, [Online]. Available: http://en.wikipedia.org/wiki/Assisted_GPS. [Accessed 5 6 2013].